# Predicting Airbnb Rental Prices
# Using Multiple Feature Modalities

Aditya Ahuja[#], Aditya Lahiri[*], Aniruddha Das[#]
adahuja@ucsd.edu[#], adlahiri@ucsd.edu[*], andas@ucsd.edu[#]

*Abstract*— **Figuring out the price of a listed Airbnb rental is an important and difficult task for both the host and the customer. For the former, it can enable them to set a reasonable price without compromising on their profits. For the customer, it helps understand the key drivers for price and also provides them with similarly priced places. This price prediction regression task can also have multiple downstream uses, such as in recommendation of similar rentals based on price. We propose to use geolocation, temporal, visual and natural language features to create a reliable and accurate price prediction algorithm.**

## I. INTRODUCTION

The pricing of Airbnb rentals is unique in the sense that they are directly set by the hosts, as compared to standardised price setting by hotel-chains. This brings in a lot of factors that need to be taken into account while deciding on the price. Price varies significantly by location, time of the year, experience of the host and amenities of the place, to name a few. Therefore, modelling this task as a machine learning problem can allow us to understand the factors that influence the rental pricing. It will also enable us to predict the prices of new rentals that are to be listed. We pose this as a regression task and use feature engineering to inform us of the price labels. We find some interesting properties during data exploration and create features to take these phenomena into account. We also try a number of machine learning algorithms to learn the price function based on these features. We interpret these results towards the end.

## II. RELATED WORK

With the rising popularity of Airbnb as a rental service in recent years, there has been a lot of work around figuring out the optimal way to price one's rental listing. The availability of open data of these listings [1] allows for machine learning modelling of these rental properties and their prices. Kalehbasti, et al. [2] use numerical and textual features to model prices for rentals in New York City in 2019. Luo, et al. [3] attempts to move beyond one city and predict rental prices for more than one city. They train on rentals from New York City and Paris, and aim for a generalization of their model on the listings from Berlin. Peng, et al. [4] use multi-modal data including textual, and geolocation data to model these rental prices. In our work, we focus on data-preprocessing and feature engineering. We use data from 8 cities in California - San Diego, Los Angeles, San Mateo, San Francisco, Santa Cruz, Pacific Grove, Oakland, Santa Monica to make the learning generalizable. We try to perform extensive feature engineering, which can aid our models to predict prices accurately, along with a variety of tuned models. We finally report the most important features, best models, their results and interpretations.

## III. DATA COLLECTION AND EDA

To generalize well across the whole of California, we decided to take Airbnb hotel-related details of eight cities spaced evenly throughout California.

### 1. Data Collection

The data for this project was collected from the official Airbnb public dataset of hotel chains in each of the eight cities - San Diego, Los Angeles, San Mateo, San Francisco, Santa Monica, Santa Cruz, Pacific Grove, Oakland. The details about *listings* consisted of information related to the host, the features associated with a particular rental listing, and the price of a rental. To make our feature space richer, we decided to incorporate the *reviews* file for a particular listing in our dataset from the site. This file contains the details about the reviews that were posted corresponding to each rental. The total size of our cumulative dataset came out to be 57,000 listings across California, along with over 200,000 reviews of all combined listings.

## 2. EDA

We begin with exploratory data analysis on our collected data. Table 1 lists a few features and the meaning of those features. Since we obtained data from multiple cities in California, we want to observe how many listings came from each city. This can be seen through the bar plot in Fig 1. Los Angeles has the most number of rentals, followed by San Diego and San Francisco. The number of people that an Airbnb rental can house is given by the accommodates feature. We see a histogram representing the number of listings corresponding to each possible value of accommodation in Fig 2. Most rentals can accommodate two people. The distribution resembles a skewed gaussian. We also plot our target variable, the price of listings. We observe that it is skewed and has outliers (*Fig 1*). We deal with this by doing a log transformation on labels as described in Section IV.

| Feature Name | Description |
|---|---|
| accommodates | Number of people the place can accommodate |
| availability_365 | Availability of listing in next 365 days |
| reviews_per_month | Number of reviews listing has |
| host_is_superhost | Is the host of the listing a superhost? |
| description | Textual description of the listing |

*Table 1.* Description of some features in the dataset

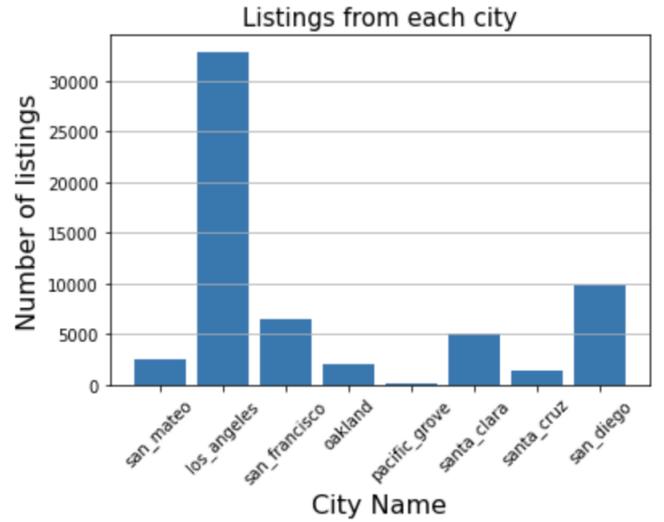

*Fig 1.* Number of listings in each city.

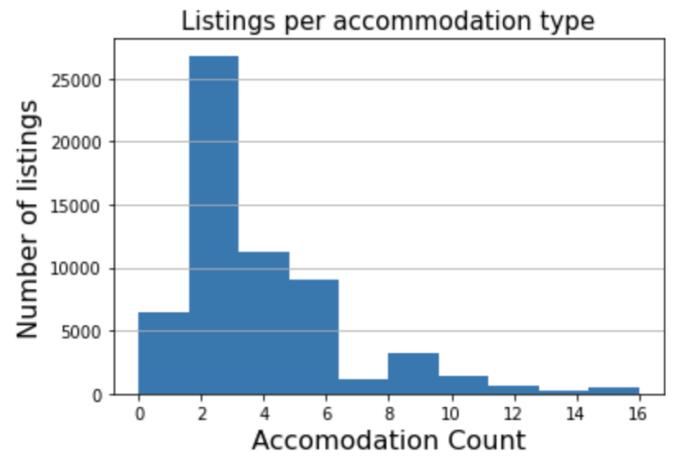

*Fig 2.* Number of listings per accommodation type.

## IV. FEATURE ENGINEERING

**1. Textual Features**

Using our *reviews* dataset for all listings in a particular city, we performed Sentiment Analysis [6] to extract the overall positive or negative sentiment score associated with a host/listing. In order to accomplish the same, we found the corresponding score for each review and set the review score for each listing by taking the mean of all review sentiments.

We also incorporated the total number of reviews associated with a listing. In addition to that, we performed TF-IDF [7] on the descriptive data of the listing to see if there were any particular keywords which are commonly associated with expensive/cheap rentals.

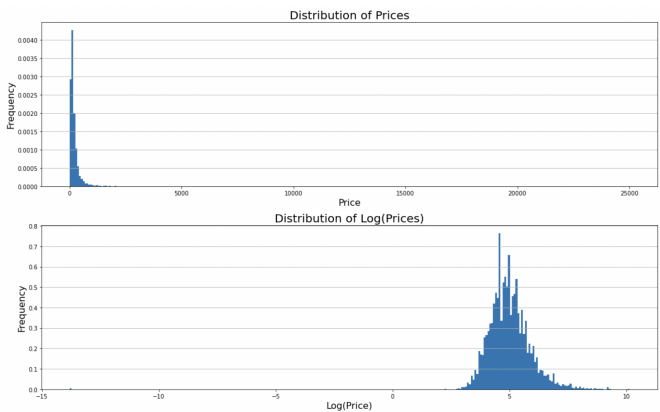

*Fig 3.* Log Transformation of Prices to deal with outliers.

We perform a log transformation on our target feature price to ensure that the prices represent a normalized gaussian distribution and to limit the effect of outliers *(Fig 1)*.

## 2. Geospatial Features

Interpretation of Longitudes and Latitudes is a difficult task, however it can add a lot of value as prices of a rental are strongly influenced by the location. In order to engineer them, we tried the following -

1. We Reverse-geocoded the latitudes and longitudes using Geopy and OpenStreetMap API to extract postal codes and rank of that area. A higher rank corresponds to the importance of the surroundings in that area - for example being near a school, hospital, or commercial area. However, due to the massive size of our dataset, we exceeded the limit on API calls. Moreover, dealing with postal codes would result in extremely huge and sparse one-hot encoded features, which would not have added much value anyway.

2. In order to reduce the number of API calls and generalize for a group of geospatial points we use Hierarchical DBSCAN with haversine as distance metric to form clusters of the listings based on the latitudes and longitudes by the centroid of these clusters. This allows us to generalize for a group of geospatial points by the centroid of these clusters. Each cluster represented 100 nearest listings. However, since all cities in our analysis belong to California, using haversine (standard metric when dealing with GIS data [5]) was redundant and euclidean seemed to perform better.

3. We optimised on the previous approach, and plotted the cluster centers on the map of California by using KMeans Clustering with Euclidean as the distance metric to form clusters of the listings based on the latitudes and longitudes.

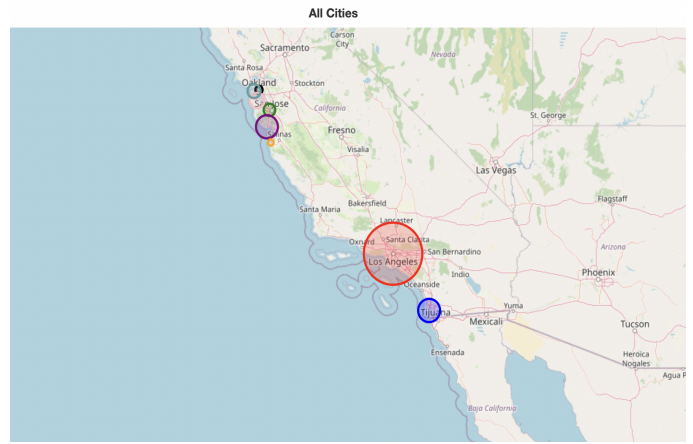

*Fig 4.* Visualising the eight cities that we do build our model on. Each circle is centered on the city center and captures the extent of a city.

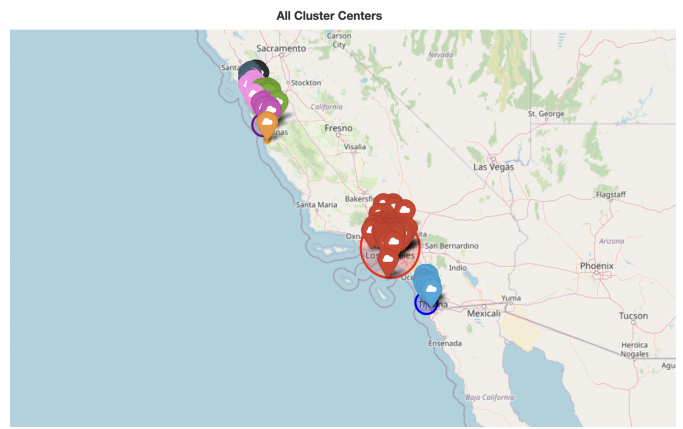

*Fig 5.* Visualising the neighbourhood clusters. Each of the 100 clusters roughly corresponds to a small set of geographically close neighbourhoods.

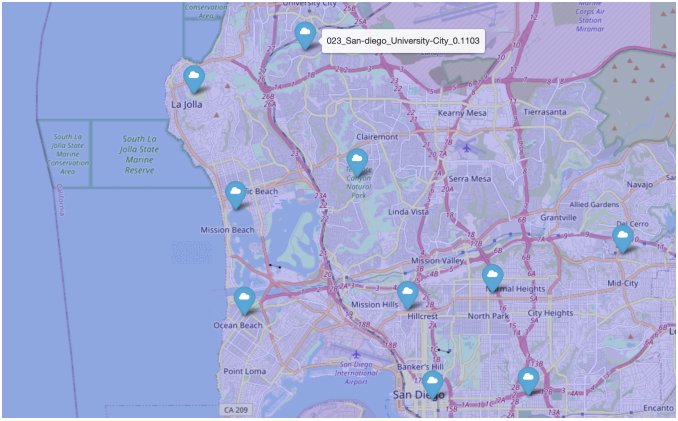

*Fig 6.* Clusters represent actual neighbourhoods. Cooking closely at the cluster centers in San Diego, it can be observed that they correspond to actually different localities - Ocean Beach, Mission Beach, Hillcrest, La Jolla, and areas around UCSD and SDSU. Thus we are able to capture features that help our model differentiate between different localities in a particular city.

In addition to the steps listed above, we used the 'neighbourhood_cleansed' feature to further group our listings based on common neighbourhoods. We one-hot encoded the above and extracted about 25 key neighbourhoods. The neighbourhood marker can have a strong influence on the average price of rentals in that area. We sorted the listings based on the number of neighbourhoods they belong to, in order to extract the most concentrated neighbourhoods with the most listings, and the most sparse ones. We represented these neighbourhoods on the map to visualize and confirm if our interpretations were valid. Moreover, we checked the popularity score of a neighbourhood to see if there are many listings in a particular neighbourhood.

### 3. Visual Features

We scraped the available images of the listing from the listing_image_url and tried to incorporate them into our feature space. The idea behind the same was that an expensive rental might have higher quality picture, better lighting, more number of available pictures, etc compared to one with a cheaper price. We used a neural network to process these images and extract the features corresponding to each listing. However, in our modelling efforts we found that this feature was not very informative. This is probably because almost all listings had great images even if the reality of the quality of listings was average. Therefore, we discard these features in creating our models.

### 4. Categorical Features

We used one-hot encoding for nominal features (including cluster labels) and label encoder for the ordinal ones. During one hot encoding of our features, we ensured not to blow up our feature space by incorporating sparseness in our dataset.

### 5. Numerical Features

We performed scaling using MinMax and Standard scaler algorithms to normalize the numerical features. We dealt with some outliers and used RobustScaler to deal with them. We also performed a log transformation on our target feature price to ensure that the prices represent a normalized gaussian distribution.

### 6. Temporal Features

The experience of a host and the new-ness factor of a listing can play a vital role in its price. We introduced these features in our dataset by allocating the number of months of experience of the host of a listing.

## V. MODELLING

After obtaining all our features, we use multiple machine learning techniques to model the price of listings using these features. We use Mean Squared Error, Mean Absolute Error and R2 score as the evaluation metrics. In Section V-1 we compare our work directly with Kalehbasti et.al [2]. In Section V-2 we list multiple models we tried and summarise the best among results obtained among them.

### 1. Comparison with existing work

We run the provided open-source code from Kalabasti et.al [2] on our dataset. We train three different model types based on models used in [2] for purpose of comparison:-

*Ridge Regression* [9] - A regularised Linear Regression model that penalises the L1 norm of weights. Ridge Regression pushes the weights to be more uniform, thereby constraining the model to be simpler.

*Gradient Boosting* [8] - A gradient boosting regressor is an ensemble algorithm with decision trees as the base learners. It provides predictions using an ensemble of weak prediction models. Each tree is trained sequentially and depends on the outputs of the trees preceding it.

*Deep Neural Network* [10] - A deep neural network is a powerful technique to model non-linear patterns in data. Deep networks have become very popular in recent years due to their ability to learn extremely well from large amounts of data.

We also use validation data to tune our model hyper parameters. By virtue of incorporating advanced geo-location based features, temporal properties, and tf-idf based description vectors, we are able to outperform their existing work. We list the Mean Squared Error (MSE), Mean Absolute Error (MAE), and R2 score on both train data and test data (10% of the dataset) in Table 1, Table 2 and Table 3. We refer to our method as "Ours". We refer to the method of Kalabasti et.al [2] as "Baseline". Train and Test are scores obtained by the respective models on training and test data. We observe that while both methods have similar performance on train data, we are able to generalise better on the test data. This is due to feature-rich representations obtained from our multi-modal features that were created during the feature engineering process. This is also reflected in the feature importances score in Fig XX. The features added by us such as the sentiment scores, tf-idf based scores, and geolocation based features have high and non-zero importances.

## 2. Top Performing Models

Since we have one hot encoded feature in our dataset, boosting algorithms can help us counter the sparseness in our dataset. While Ridge Regression, deep neural networks performed well on the training and test dataset, we observed that gradient boosting with minimal hyper parameter tuning out performed the above two.

|  | Ours | | Baseline | |
| --- | --- | --- | --- | --- |
|  | Train | Test | Train | Test |
| Boosting | 0.09 | **0.16** | 0.10 | 0.32 |
| Ridge | 0.21 | **0.20** | 0.21 | 0.27 |
| Deep NN | 0.12 | **0.17** | 0.15 | 0.23 |

*Table 2.* Mean Squared Errors comparison between our method and the baseline for different implemented models.

|  | Ours | | Baseline | |
| --- | --- | --- | --- | --- |
|  | Train | Test | Train | Test |
| Boosting | 0.15 | **0.27** | 0.17 | 0.42 |
| Ridge | 0.32 | **0.31** | 0.32 | 0.37 |
| Deep NN | 0.25 | **0.29** | 0.28 | 0.34 |

*Table 3.* Mean Absolute Errors comparison between our method and the baseline for different implemented models.

|  | Ours | | Baseline | |
| --- | --- | --- | --- | --- |
|  | Train | Test | Train | Test |
| Boosting | 0.8 | **0.6** | 0.77 | -0.51 |
| Ridge | 0.56 | 0.58 | 0.56 | **0.63** |
| Deep NN | 0.79 | **0.70** | 0.72 | 0.65 |

*Table 4.* R2 score comparison between our method and the baseline for different implemented models.

Specifically, we tried two more boosting algorithms-

*LightGBM* [11] - It is a gradient boosting framework that uses tree-based learning algorithms. It is designed to be distributed and efficient in terms of faster training speed, lower memory usage, and thus allows our model to handle large scale data.

*XGBoost* [12] - It is a gradient boosting algorithm which utilized computational tricks to make the model more robust and scalable. It uses a penalty function similar to an elastic net which allows it to incorporate sparse modelling too.

We used GridSearchCV to further optimize the model parameters. We found that LightGBM performed the best with the following results (summarised in *Table5*) - We obtained these results with tuned hyperparameters of number of estimators set to 1000, learning rate of 0.1 and alpha regularisation of 0.5 on our LightGBM model.

|     | XGBoost | | LightGBM | |
| --- | --- | --- | --- | --- |
|     | Train | Test | Train | Test |
| MSE | 0.10 | 0.14 | 0.06 | **0.13** |
| MAE | 0.23 | 0.26 | 0.17 | **0.24** |
| R2  | 0.81 | 0.73 | 0.89 | **0.76** |

*Table 5.* Summary of results obtained with other models.

The reason why LightGBM performed the best is because of a key difference in the way it is trained. It uses the best leaf-wise split to build the tree instead of depth-wise or level-wise splits. This leads to better performance compared to vanilla gradient boosting on our dataset.

## VI. RESULTS AND INTERPRETATION

Our best model LightGBM uses features like latitude, description score, longitude, review sentiments, neighbourhood popularity, experience of the host, availability of the listing, etc to fit the training data. Most of the features which were engineered by us played a crucial role in predicting the price of rentals. The high R2 score achieved helps us conclude that these features were indeed able to explain the variance of our model. In *Fig5* we can see the feature importances listed in a sorted fashion.

Interpretation of our top 6 features *Fig5*:

1. Latitude: The cities we have selected in California were indeed at different latitude levels. It is a well known fact that rental prices vary greatly based on the city it is based in. The longitude also helps but the latitude being the most important feature used by our model does make sense.
2. Description Score: Extracting keywords from the description of a listing was done with the intuition that expensive places usually have certain keywords consistent with their description, and vice versa. Our model confirms that this assumption of ours while feature engineering was true.
3. Sentiment Scores: Cumulative review scores of a listing can heavily influence the pricing of the rental. An expensive listing would usually have a higher positive review score, and vice versa. Our model does capture the feature, and hence approves the assumption.
4. Neighbourhood Popularity: Neighbourhoods having a higher concentration of listings would usually influence the price of rentals across the region.
5. Host Since: A new host may not be that reliable as compared to a host with many years of experience. This might be even further correlated to the quality of service in that particular listing. Thus, experience should and does play a vital role in the price of a rental.

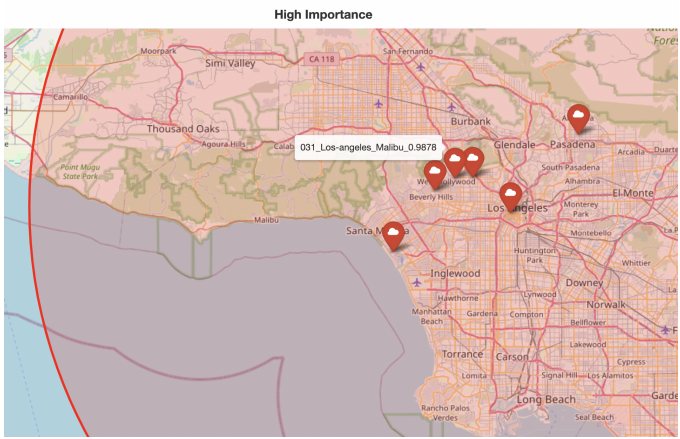

*Fig 7*. Most informative Neighbourhoods. Six out of the seven most informative neighbourhoods were in Los Angeles (the other was in the center of San Francisco). It can thus be seen that for localities with a high cost of living (Hollywood Hills, Malibu, Beverly Hills), it is informative to add cluster features to help inform the model about this relative pricing bias.

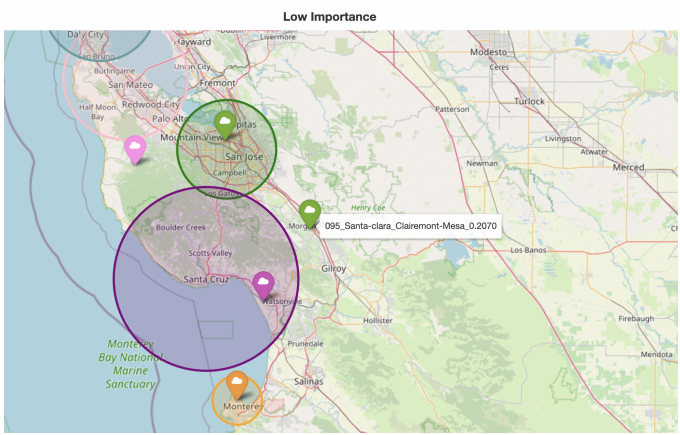

*Fig 8.* Least Informative Neighborhoods. It can be seen that neighbourhood clusters that lie outside city boundaries (outskirts) have low importance when predicting prices. For instance, in the above plot the "Clairemont Mesa" neighbourhood has low importance which might be related to it being on the outskirts.

## VII. CONCLUSION & FUTURE WORK

In our work, we focus on predicting rental prices for listings obtained from various cities inside the state of California using multi-modal features. Our methods are generalisable and can be extended for listings from any state or country. Future work can revolve around trying to see how our features and models perform in listings from other states. Furthermore, we can also add some state and location specific features to improve the model. For instance, we can add San Diego specific features such as marking the La Jolla neighborhood as expensive. These domain specific features can help improve the model. The trade-off here is that such knowledge is not easily obtained and non-generalisable.

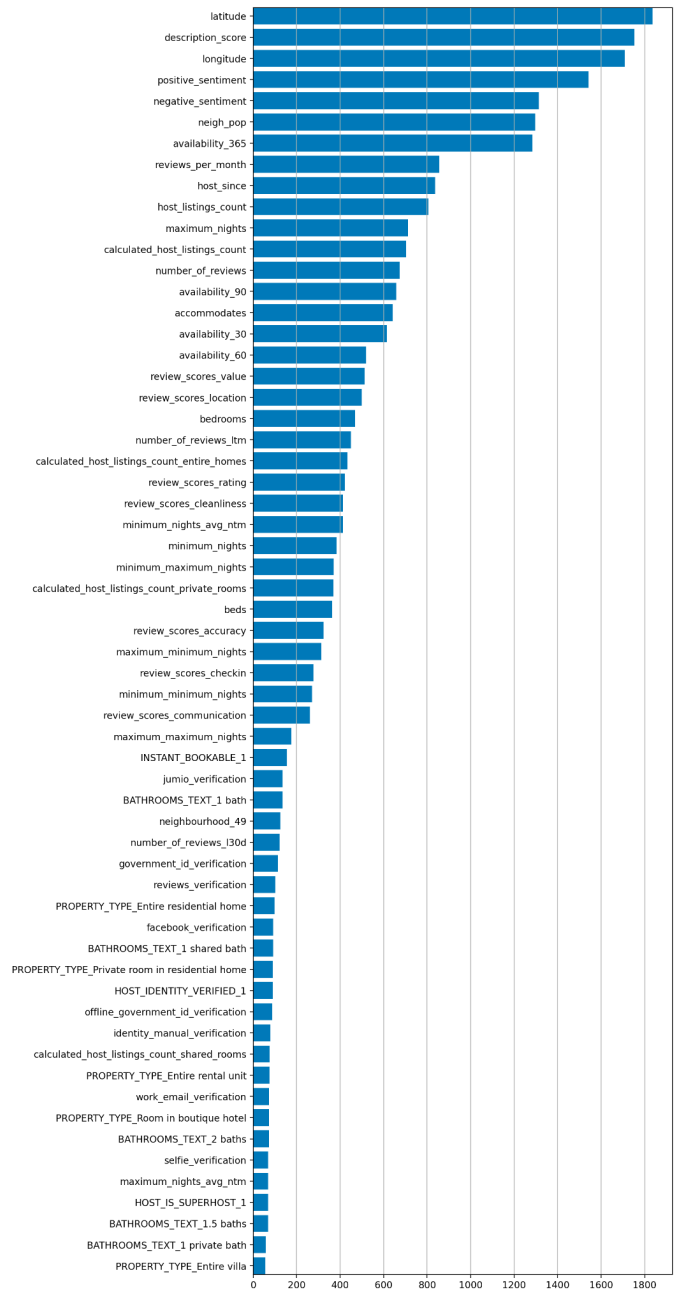

*Fig 9*. Feature Importance of the top 60 most important features used in our model.